\documentclass[letter, 10pt, conference]{ieeeconf}
\IEEEoverridecommandlockouts                              
\usepackage{cite}
\usepackage{amsmath}
\usepackage[utf8]{inputenc}
\usepackage[english]{babel}
\usepackage[T1]{fontenc}
\usepackage{soul}
\usepackage{color}
\usepackage{graphicx}
\usepackage{xcolor}
\usepackage{subcaption}

\begin{document}
\bstctlcite{IEEEexample:BSTcontrol}
\title{\LARGE 
Combining Planning and Learning of Behavior Trees for Robotic Assembly}

\author{Jonathan Styrud\authorrefmark{1}\authorrefmark{3}, Matteo Iovino\authorrefmark{2}\authorrefmark{3}, Mikael Norrlöf\authorrefmark{1}, Mårten Björkman\authorrefmark{3} and Christian Smith\authorrefmark{3}
\thanks{This project is financially supported by the Swedish Foundation for Strategic Research and by the Wallenberg AI, Autonomous Systems, and Software Program (WASP) funded by the Knut and Alice Wallenberg Foundation. The authors gratefully acknowledge this support.}
\thanks{\authorrefmark{1}ABB Robotics, Västerås, Sweden}
\thanks{\authorrefmark{2}ABB Corporate Research, Västerås, Sweden}
\thanks{\authorrefmark{3}Division of Robotics, Perception and Learning, Royal Institute of Technology (KTH), Stockholm, Sweden}}

\maketitle

\begin{abstract}
Industrial robots can solve very complex tasks in controlled environments, but modern applications require robots able to operate in unpredictable surroundings as well. An increasingly popular reactive policy architecture in robotics is Behavior Trees but as with other architectures, programming time still drives cost and limits flexibility. There are two main branches of algorithms to generate policies automatically, automated planning and machine learning, both with their own drawbacks. We propose a method for generating Behavior Trees using a Genetic Programming algorithm and combining the two branches by taking the result of an automated planner and inserting it into the population. Experimental results confirm that the proposed method of combining planning and learning performs well on a variety of robotic assembly problems and outperforms both of the base methods used separately. We also show that this type of high level learning of Behavior Trees can be transferred to a real system without further training.
\end{abstract}


\section{Introduction}
Today, industrial robots can solve very complex tasks in controlled environments, but modern industrial applications in workspaces shared with humans, require robots able to operate in unpredictable surroundings as well. One common way to deal with this is to control the robot with a reactive policy such as Behavior Trees (BTs)~\cite{colledanchise_behavior_2018,iovino_survey_2020}. Also, with increased variation and smaller production batches, the time needed for programming, system integration, validation and verification is prohibitively long. It is therefore desirable that new robot policies or programs can be created automatically.
\par There are two main branches of algorithms to generate policies automatically, both with their own drawbacks. The first branch, automated planners~\cite{ghallab_automated_2016}, can be very efficient but is limited by the amount of knowledge that has been given to the planner in advance, as planners typically do not interact with the environment. Also, beyond a certain problem complexity, planners tend not to scale well and cannot always find a solution within reasonable time. The second branch, machine learning algorithms, often do interact with the environment and is thus not limited by a static set of knowledge. They can also scale  well for complex problems, using a probabilistic rather than exhaustive approach. However, for many problems of low to medium complexity, an automated planner can be many orders of magnitude faster. Another drawback is that many of the most efficient machine learning algorithms, often based on Reinforcement Learning, are adapted to neural networks that have several important disadvantages compared to BTs, as neural networks are not particularly transparent or as modular.
\par In this paper, we propose a method to combine the two branches by taking the result of an automated planner and inserting it into the population of a Genetic Programming (GP) algorithm, a common machine learning algorithm. In this way, the  learning algorithm converges much faster than if starting from scratch, and is still able to solve problems the automated planner cannot. For the evaluation of the machine learning algorithm to be feasible we run the training in a controlled simulated environment, showing that convergence is indeed faster. This would often lead to problems when running the resulting policy in the real world. We mitigate this by using high level behaviors and eventually demonstrate that it works on a real robot without any extra training.


\section{Background and related work}
In this section we provide some background on the two key technologies, Genetic Programming and Behavior Trees, and summarize the related work.

\subsection{Genetic Programming}
\begin{figure}[htbp]
\centerline{\includegraphics[width=0.45\textwidth, height=8cm]{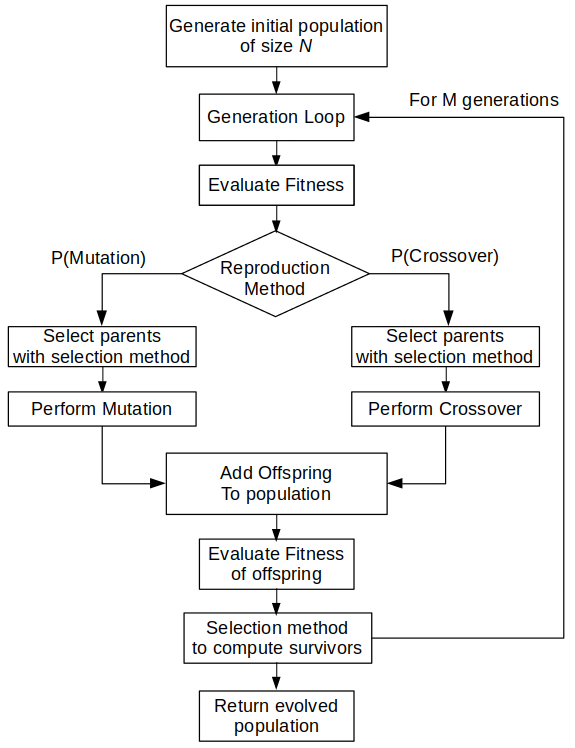}}
\caption{Scheme of GP execution flow.}
\label{fig:GP_scheme}
\end{figure}

Genetic Programming is an optimization algorithm that can evolve programs represented as trees~\cite{koza_genetic_1992,sloss_2019_2020}; see Figure \ref{fig:GP_scheme} for a schematic overview. Populations of individuals generate offspring through the functions of \emph{crossover} and \emph{mutation}. A selection mechanism decides which individuals of the population to keep. Examples of selection mechanisms are \emph{elitism} (keeping the $n$ highest ranking), \emph{tournament selection} (individuals are compared pair-wise), \emph{rank selection} (the probability to keep an individual is proportional to its rank in the population). The survival selection is based on a fitness function that assigns a score to each individual based on how well it performs in solving the task. There are many variations of GP, e.g. where a grammar is defined and programs are generated from simple lists of integers (Grammatical Evolution), or where only programs parameters are allowed to change and the genotype is represented as fixed-length strings (Genetic Algorithm).

\subsection{Behavior Trees}
Behavior Trees originated in the gaming industry as an alternative architecture to Finite State Machines (FSM) to represent policies~\cite{colledanchise_behavior_2018}. BTs have explicit support for task hierarchy, action sequencing and reactivity, and improve on FSMs especially in terms of modularity and reusability~\cite{iovino_survey_2020}. 
\par
In a BT, the branching nodes are called \emph{control nodes}, (hexagons and rectangles in the example Figure \ref{fig:blocking_bt_planned}), while the leaves are called \emph{execution nodes}, \emph{leaf nodes} or simply \emph{behaviors} (ovals). During execution, \textit{tick} signals are propagated from the root down the tree at a specified frequency. There are several different variants of control nodes, the two most common being \emph{Sequence} nodes and \emph{Fallback} (or \emph{Selector}) nodes. Sequence nodes execute their children in a sequence, returning once all succeed or one fails. Fallback nodes also execute their children in a sequence but return when one succeeds or all fail. Execution nodes executes a behavior when ticked and returns one of the status signals \textit{Running}, \textit{Success} or \textit{Failure}. They are often categorically split into Action nodes and Condition nodes where the latter differ in that they always finish immediately with \textit{Success} or \textit{Failure}, typically checking the status of some data or sensor. BTs are closely related to decision trees with the main differences being that BTs can hold a \textit{Running} state so that tasks can execute for longer than one tick. The return state \textit{Running} is also crucial for reactivity, allowing other actions to preempt non-finished actions. For more detail on behavior trees, see e.g.~\cite{colledanchise_behavior_2018}.
\par
The modularity of a BT is particularly relevant when using evolutionary algorithms such as GP, because any subtree can be added to the gene pool and be re-used in following generations. With genetic crossover, parents generate offspring by exchanging gene sequences and in BTs this can be done with subtrees without compromising the logical execution of the tree~\cite{colledanchise_learning_2019} and the safety guarantees~\cite{colledanchise_how_2017}. 

\subsection{Related work}
The potential for combining planning and learning has been recognized by many and in particular various combinations of planning and Reinforcement Learning have been proposed and used; see for example~\cite{grounds_combining_2008,faust_prm-rl_2018,francois-lavet_combined_2019,moerland_model-based_2020}. However, the challenges of creating neural networks are quite different from those of creating BTs.
\par
In more closely related work, BTs are instead generated with evolutionary methods such as GP. Notably, a number of studies have shown that solutions can be found that outperform manual designs~\cite{hutchison_evolving_2010,scheper_behavior_2015,jones_evolving_2018,perez_evolving_2011,colledanchise_learning_2019}. In~\cite{scheper_behavior_2015} the authors evolved BTs to control an Unmanned Aerial Vehicle while~\cite{jones_evolving_2018}, used the BTs to control a swarm of Kilobots executing a foraging task. As BTs originated in the gaming industry, video game benchmarks are more common than robotics. As an example~\cite{perez_evolving_2011,nicolau_evolutionary_2017,colledanchise_learning_2019} used the Mario AI environment as a benchmark while~\cite{hutchison_evolving_2010} had the BT agent play the game DEFCON and~\cite{mcclarron_effect_2016} a version of Pac-Man. 
\par 
Looking in more detail at the algorithms used, \cite{perez_evolving_2011} and \cite{nicolau_evolutionary_2017} use a different evolutionary method than GP called Grammatical Evolution (GE). In~\cite{nicolau_evolutionary_2017}, the evolutionary method is combined with a planner as the GE algorithm is used together with an A* search algorithm. GE in general suffers from a number of drawbacks, as the abstraction between genotype and phenotype can make it more difficult to analyze and implement heuristics and negatively affects locality~\cite{rothlauf_locality_2006}. \cite{colledanchise_learning_2019} use GP to some extent but their method is largely based on a one-step search from the current best solution, only using GP as a fallback when the one-step search can not find an improvement. \cite{scheper_behavior_2015,mcclarron_effect_2016,jones_evolving_2018} use variants of GP that are somewhat different from the algorithm used in our experiments as specified in Section \ref{section:gp}. The algorithm in~\cite{hutchison_evolving_2010} is not named but it closely resembles a GP algorithm.
In our previous work we used a version of GP very similar to the one described later in this paper to learn robotic manipulation tasks~\cite{iovino_learning_2020}.

Also related is work on automated planners. In this paper we use an adaptation of the Planning Domain Definition Language (PDDL)-style planner from~\cite{colledanchise_towards_2019} where the authors leverage the idea of backchaining by starting from the goal conditions and proceeding backwards, iteratively finding the actions that fulfill the necessary conditions. In~\cite{tumova_maximally_2014} the action capabilities of a NAO robot tasked with grasping and transporting objects were modeled as a transition system and the goal as a State-Event Linear Temporal Logic (LTL) formula. LTL is also used in~\cite{colledanchise_synthesis_2017} to synthesize a BT for a maze navigation task. The rigorous and complex formulations of the LTL formula make the approach scale poorly for more complicated tasks or less predictable environments. In~\cite{holzl_reasoning_2015}, an Extended Behavior Tree (XBT) structure was proposed that uses Hierarchical-Task-Network (HTN) planning. \cite{rovida_extended_2017} suggested a different extended Behavior Tree (eBT), using concepts from HTN with a PDDL-planner, to solve a robot kitting task. Another PDDL-style extension was proposed in~\cite{giunchiglia_conditional_2019}, called Conditional Behavior Trees (CBT) and was demonstrated for a robotic manipulation task.
Please see Sections 4.1.2 and 2.4 in~\cite{iovino_survey_2020} for a more exhaustive list of related work.

\section{Proposed method}
\subsection{Automated planner}
The method of this paper will be valid for any planner that can generate a BT that solves at least some part of the problem. For our experiments we implemented a simple PDDL-planner based on~\cite{colledanchise_towards_2019}. Here, all behaviors are manually designated a set of pre-conditions that must be fulfilled in order to execute the behavior successfully and a set of post-conditions that will be fulfilled after the behavior is successfully executed. A set of goal-conditions is also specified that defines the robot's task. The planner then expands the tree iteratively until all conditions are met. The complete code of the planner and all other algorithms are available online\footnote{https://github.com/jstyrud/planning-and-learning}.

\subsection{Genetic programming}\label{section:gp}
\par Our GP algorithm is mostly standard \cite{koza_genetic_1992} but with some specific adaptations to fit the generating of BTs. Figure~\ref{fig:GP_scheme} provides an overview of the implementation. One adaptation in the algorithm is to limit the structure of the tree to guarantee a valid and minimal BT. Some of these limitations were inspired by \cite{mcclarron_effect_2016} but we differ by allowing any control node type to be the root, and not implementing constraints on the ordering of execution nodes. Using all the constraints could have sped up the learning, but the limit of possible tree types may generalize worse to a wider range of tasks.
\par The main constraints that are implemented are firstly that two identical condition nodes cannot be placed next to each other as this would be redundant. For the same reason, a control node may not have an identical control node as a parent. Also, only trees where all control nodes have at least one child are allowed. 
\par While designing the possible changes that the algorithm can do to the BTs, some guiding principles were to make the set complete so that any valid tree can be generated given enough time and to make sure that improving changes can be done with as few steps as possible while keeping the set minimal. This ensures that time is not wasted on changes that do not improve the performance.
\par When it comes to mutating the BTs, the possible mutations are \emph{adding nodes}, \emph{deleting nodes} and \emph{changing nodes}. When adding a random node, a random control node is selected with 50\% probability and a random execution node otherwise. This probability reflects the observed distribution between the node types in typical trees. When the mutation changes a node into another node, the same probabilities are used for selecting the new node. Some mutations mandate more immediate steps to ensure that the BT constraints are fulfilled. For example when adding a new control node, two random children are added. This makes the overall progress faster as time is not wasted on invalid BTs.
\par For crossover, two individuals are used and a random subtree from each individual is selected. A subtree can be anything from a control node and all its descendants to single execution nodes. These subtrees are inserted into the other individual at a random point. When doing crossover the subtrees are traditionally swapped but we have found this to be much less efficient. We believe that this is because typically a given individual will not have many subtrees that it would do better without and the conventional crossover would need the algorithm to first add a random dummy subtree only for it to be swapped in a later step making it much less likely to find a fitness improvement. Note that this heuristic affects learning speed, but not which parts of the solution space that are reachable.
\par The design of a fitness function to evaluate and rank individuals is crucial for performance and is often non-trivial. The function should have a slope in the solution space that is as smooth as possible towards the optimal solution for the algorithm to follow. It is also valuable if the fitness function is general, and not too connected to a specific task, so that it can be easily re-usable. In our experiments the tasks consist of moving objects, a natural fitness is therefore based on the Euclidean distance from the objects to their goal positions. Because there is an uncertainty and variability in the object positions both in simulation and reality, up to $\delta$~mm is removed from each distance so that trees that are all solving the task get the same fitness score. Otherwise the algorithm will generate numerous BTs that only vary in not meaningful ways only to keep the ones where the random variations made the errors somewhat smaller. A penalty is also added for the number of nodes in the tree in order to make them as small as possible. It is important that this penalty is small enough so that adding meaningful nodes still improves the fitness, while being large enough to drive the trees smaller. A penalty is added corresponding to $\lambda$~mm per node in the tree. Finally, the equivalent of $\tau$~mm penalty is added to any tree that ends by timeout, to avoid trees that "get lucky" when interrupted because of running out of time and $\phi$ is added to trees that end in Failure state. The complete equation for the fitness that the algorithm maximizes becomes
\begin{equation}
    \mathcal{F} = - \sum_\mathcal{O} max(0, ||o - g|| - \delta) - \lambda L - \tau T - \phi F
\label{eq:fitness}
\end{equation}
where $\mathcal{O}$ is the set of objects with current positions $o$ and goal positions $g$, $L$ is the number of nodes and $T$ is 1 if the tree ended by timeout and 0 otherwise and $F$ is 1 if the tree ended if Failure state and 0 otherwise. See Table \ref{GP pars} for a list of parameter values used in the experiments in Section~\ref{sec:exper}
for the most important parameters of the GP algorithm.

\begin{table}[htbp]
\tiny
\scriptsize
\caption{GP parameters}
\begin{center}
\begin{tabular}{|c|c|}
\hline
\bf{Parameter description} & {\bf Value} \cr
 
\hline
Population size & 16 \cr
\hline
Initial random BT size & 8 \cr
\hline
Parents selected for mutation & 8 \cr
\hline
Number of mutation offspring per parent & 2 \cr
\hline
Mutation probabilities for add, delete, change & 40\%, 30\% and 30\%\cr
\hline
Parents selected for crossover & 8 \cr
\hline
Number of mutation offspring per parent & 2 \cr
\hline
Selection method (reproduction and survival) & Rank proportional selection \cr
\hline
Number of elites & 2 \cr
\hline
Error allowed for each target ($\delta$) & 0.4mm \cr
\hline
Length penalty for each node ($\lambda$) & 0.1 \cr
\hline
Penalty for ending by timeout ($\tau$) & 10 \cr
\hline
Penalty for ending in Failure state ($\phi$) & 50 \cr
\hline
\end{tabular}
\end{center}
\label{GP pars}
\end{table}

\subsection{Combining planning and GP}
In order to use the results from the planner to speed up the GP, the resulting BT from the planner is inserted into the starting population. However, this is not enough as it is possible for the algorithm to discard the planned BT from the population in a subsequent generation if it has low fitness, even if the tree could be usable. This could be when one of the early nodes fails, and the entire tree will fail before executing the later, perhaps useful, parts and thereby getting a low fitness score. Therefore, the GP is forced to always keep the planned BT in the population. Further, even with the planned BT in the population, if it has a low fitness score it will also have a low probability of being selected for crossover and mutation. To counter this, two additional versions of the algorithm have been implemented and tested where the fitness of the planned BT is temporarily boosted during parent selection to equal the highest fitness of the rest of the population. For the first of these versions, the fitness is boosted only for crossover parent selection and for the second version the fitness is boosted for both crossover and mutation parent selection.


\section{Experiments}
\label{sec:exper}
Experiments of four different assembly tasks with LEGO DUPLO bricks were performed. In all cases we ran the same algorithm with identical parameters to demonstrate its generality. We obtained results for various setups with and without a planner baseline to compare the learning curves. The parameters for the GP setup for all experiments are given by Table \ref{GP pars}. The BTs are implemented using the PyTrees framework\footnote{https://github.com/splintered-reality/py\_trees, version 2.0.16} and every BT was run in a simulated AGX Dynamics\footnote{https://www.algoryx.se/agx-dynamics/, version 2.29.3.1} environment to obtain the fitness. In the simulation, all manipulation is done by a two-finger prismatic gripper and simulated friction grips. Since the simulation is deterministic, every BT was only run once and the result stored in a hash table in case the same BT would occur in a later generation, for example if it was one of the survivors. For all experiments, the only control nodes used are Fallback and Sequence nodes without memory. That is, the nodes will tick any previously successful child again, making the tree more reactive but forcing the use of condition nodes to block unnecessary rerunning of some behaviors.
\par Each tree is allowed at most 200 ticks to solve the task, corresponding to 20s simulation time at the 10Hz tick-rate used. They are also allowed a maximum of 30s real time. Further, the episode is  aborted if the root returns Failure or if it returns Success twice in a row. The reason for requiring two successive success states is that some trees do not recognize that the task is completed and would start executing behaviors again after the first Success if given the chance, and such trees should be avoided. 

\subsection{Behaviors}
The set of available behaviors was varied slightly between the tasks in the experiments, but all were constructed from the templates described here. Behaviors whose name ends with "?" are conditions that never return Running while behavior with "!" are actions that can return Running. Actions will return Failure if not completed within 50 ticks or if some other error occurs such as trying to pick a brick when something is already picked.
\newline\textbf{picked a?} returns Success if \textit{brick a} is grasped and held by the gripper, Failure otherwise.
\newline\textbf{a at pos p?} returns Success if \textit{brick a} is within 1mm of \textit{position p}, Failure otherwise.
\newline\textbf{a on b?} returns Success if $x$ and $y$ coordinates of \textit{brick a} are within 1mm of $x$ and $y$ coordinates of \textit{brick b} and \textit{brick a} is above \textit{brick b} but not more than 10mm above, Failure otherwise.
\newline\textbf{pick a!} picks up \textit{brick a} through a series of moves and using the gripper. Fails if another brick is already picked.
\newline\textbf{place on a!} attempts to place whatever brick is in the gripper on top of \textit{brick a}. Fails if the gripper is empty.
\newline\textbf{place at pos p!} attempts to place whatever brick is in the gripper at \textit{position p}. Fails if the gripper is empty.
\newline\textbf{put a on b!} combines pick and place into one action and places \textit{brick a} on \textit{brick b}. This allows for a smaller tree and for the behavior to return Success immediately if \textit{brick a} is already on \textit{brick b}, eliminating the need for guarding conditions. Fails if another brick is already picked.
\newline\textbf{put a at pos p!} combines pick and place as above but places at \textit{position p} instead of on a brick.
\newline\textbf{apply force a!} moves to the top of \textit{brick a} and presses downward on the brick with 1N for 0.2s. Fails if the gripper is not empty.

\subsection{Evaluation tasks}

Four simulated tasks were defined to evaluate alternative ways to combine BT and GP for more efficient tree search.

\begin{figure}[htbp]
\centerline{\includegraphics[width=0.5\textwidth]{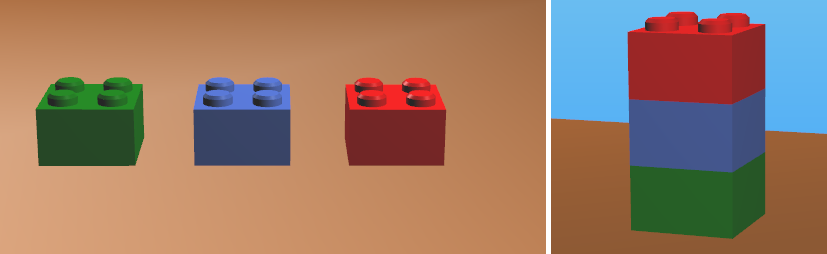}}
\caption{Start (left) and goal (right) configurations of task 1.}
\label{fig:tower}
\end{figure}

\begin{figure}[htbp]
\centerline{\includegraphics[width=0.5\textwidth]{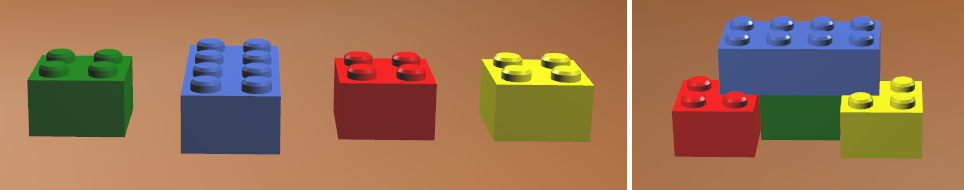}}
\caption{Start (left) and goal (right) configurations of task 2.}
\label{fig:cro}
\end{figure}

\paragraph{Task 1}
The first task was to stack three bricks on top of each other in a certain order and to press them together so that they are properly fitted. This was handled well by the planner with the information it had available. The start and goal configurations can be seen in Figure \ref{fig:tower}. This task was run for 200 generations for all variants. The complete set size of behaviors used for this task was 22 and the \emph{put} behaviors were not used. 
\paragraph{Task 2}
The second task was a structure where three bricks were to be placed on the table in a U-shape and a fourth longer brick was to be placed on top of the middle brick. The start and goal configurations of the second task can be seen in Figure \ref{fig:cro}. The planner does not know that all the bricks on the table needs to be placed before the longer brick on top to avoid collision. The particular task may be regarded as somewhat contrived but the results generalize to any task where the planner is lacking some information about physics or something else that a simulation or real world experiment will reveal. As this task is somewhat more complicated it was run for 1000 generations for the unboosted variants and 500 generations for the boosted variants. The complete set size of behaviors used for this task was 35 different behaviors and the \emph{put} behaviors were not used.

\begin{figure}[htbp]
\centerline{\includegraphics[width=0.5\textwidth]{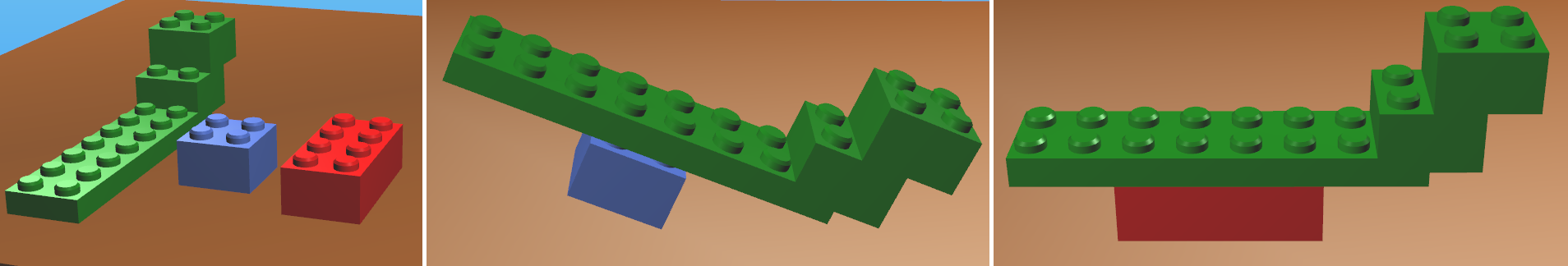}}
\caption{Start (left), unbalanced (middle) and a configuration that satisfies goal constraints (right) of task 3.}
\label{fig:balance}
\end{figure}

\paragraph{Task 3}
In the third task, the goal was to elevate a large brick with uneven balance to the height of another brick. Because of the uneven balance, a wider base must be used. The planner was unaware of physics and balance and thus by chance planned a tree where the stacked bricks fall over. The start and goal configurations of task 3 can be seen in Figure \ref{fig:balance}, as well as the result of an unbalanced attempt from the planner. To solve this task, the GP needs to do more than simply reordering subtrees so the composite \emph{put} behaviors were used to limit the search space and make experiment runs possible in reasonable time. In early tests, the GP was able to find solutions where the gripper simply held the bricks in order to keep them from falling over. To avoid such solutions we added another term to the fitness function $\mathcal{F}$  in Eq.~\ref{eq:fitness} for this particular task by introducing a penalty $\eta$ if the gripper was holding any of the bricks at the end of the episode. The fitness function for this task then becomes
\begin{equation}
    \mathcal{F'} = \mathcal{F} - \eta H
\label{eq:fitnesstask3}
\end{equation}
with $\eta=100$ and H is 1 if the gripper is holding a brick at the end of the episode and 0 otherwise.
The GP was run for 800 generations for all setups. The complete set size of behaviors used for this task was 23 different behaviors.

\begin{figure}[htbp]
\centerline{\includegraphics[width=0.5\textwidth]{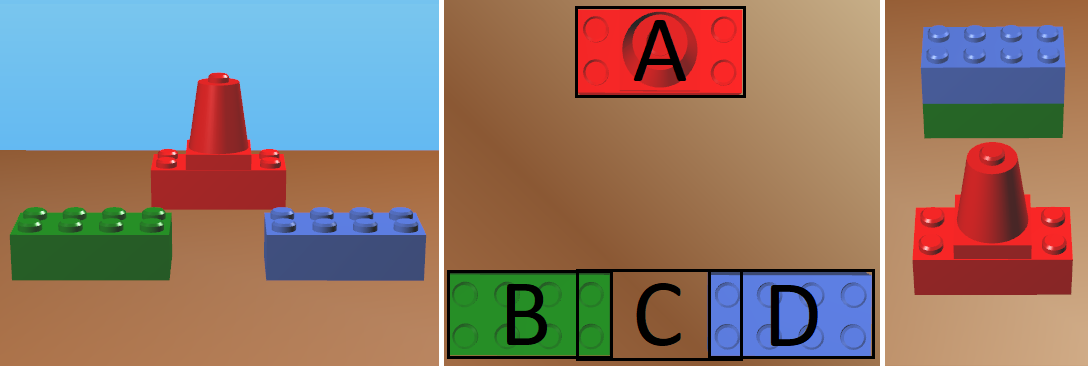}}
\caption{Start (left) and goal (right) configurations of task 4. Labeled table positions are shown in the middle picture.}
\label{fig:blocking}
\end{figure}

\paragraph{Task 4}
The fourth task has similar properties to the famous Tower of Hanoi puzzle where the bricks are limited in where they may be put and are blocking each other, enforcing certain orders of moves. The start and goal configurations of task 4 can be seen in Figure \ref{fig:blocking}. The only allowed table positions are the ones labeled A, B, C, and D in the figure. The green and blue bricks should be placed on position A, that is blocked by the red brick, and the red brick should be placed in position C, blocked by green and blue. The red brick cannot be stacked with the other bricks, as illustrated and ensured by the red cone on top. One solution is to stack green and blue on B, enabling red to be moved to D, and then moving green and blue to A before finally moving red to C. The task is especially interesting because it requires planning several steps ahead to get out of local minima. As in task 2, the planner was unaware of possible collisions and planned a naive tree that failed. Again, the GP needs to do more than simply reordering subtrees so the composite \emph{put} behaviors were used. The GP was run for 300 generations for all setups. The complete set size of behaviors used for this task was 27 different behaviors.


\section{Results}
On the four tasks we compare the results from the planner (dotted lines) to those of the GP algorithm with (blue) and without (red) the planned baseline solution as input. We also test the two versions for which the baseline is boosted, when boosting is done only for crossover parent selection (green) and for parent selection for both mutation and crossover (black). We also plot a dashed line representing the best found solution.
\par
\begin{figure}[htbp]
\centerline{\includegraphics[width=0.5\textwidth]{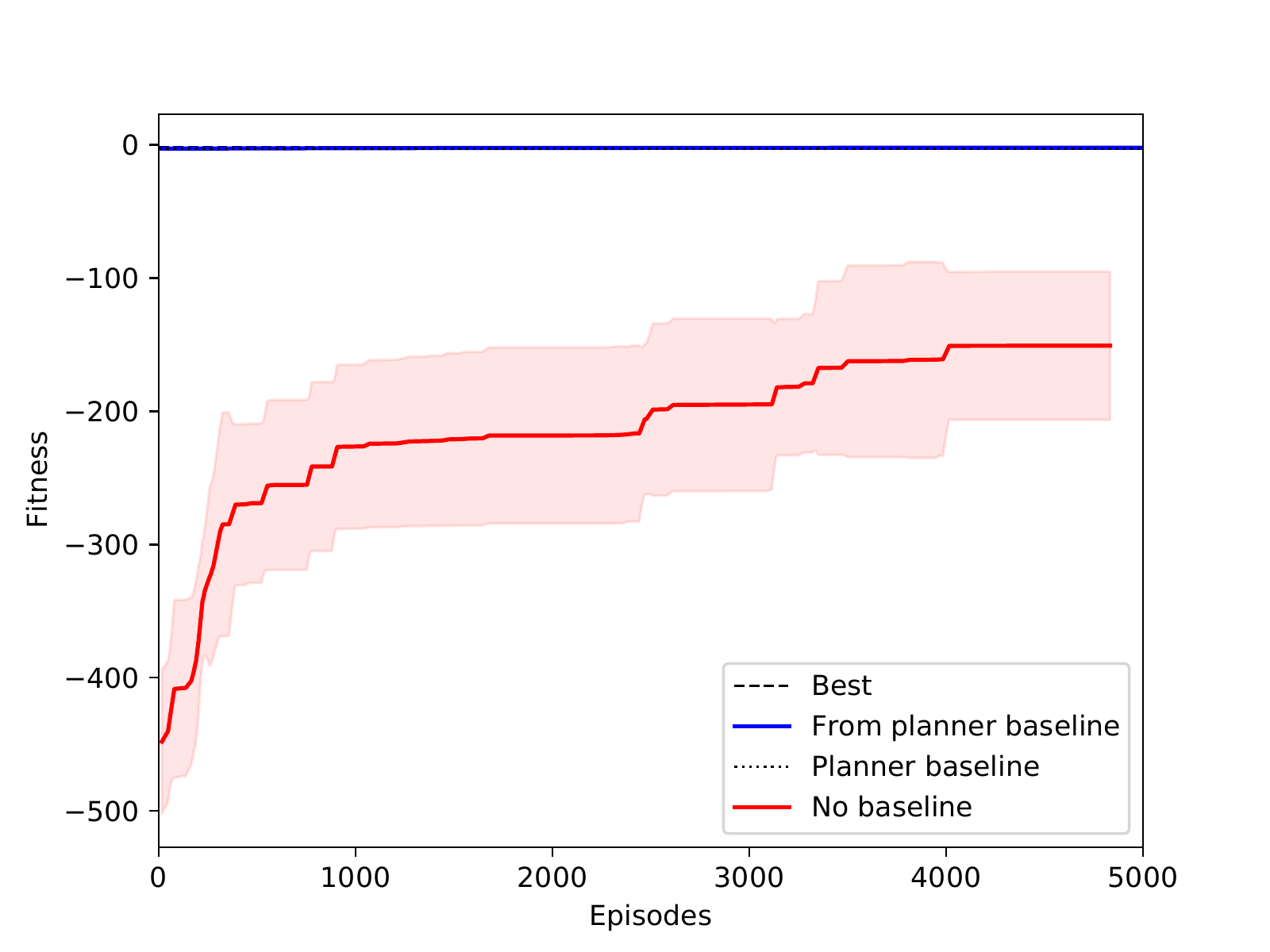}}
\caption{Learning curves for task 1.}
\label{fig:tower_learning_curve}
\end{figure}

Figure \ref{fig:tower_learning_curve} shows the learning curves for task 1 with and without a planner baseline. The diagram shows the mean of 10 runs for each variant with different random seeds and shaded standard deviation. We only show mean of the best individuals of each generation for the different runs, not the mean of the whole population. Because of the use of elitism, each generation can only improve, never degrade. The $x$-axis shows the number of episodes, meaning the number of unique simulations, since this is what drives computation time, not the number of generations. With the parameters in Table \ref{GP pars}, each generation could generate anything between 0 and 32 new unique simulations as any particular BT will only be run once. Since the planner already solves task 1, the curve from the planner baseline moves only slightly as the GP finds a somewhat smaller tree to do the same task.
\par
\begin{figure}[htbp]
\centerline{\includegraphics[width=0.5\textwidth]{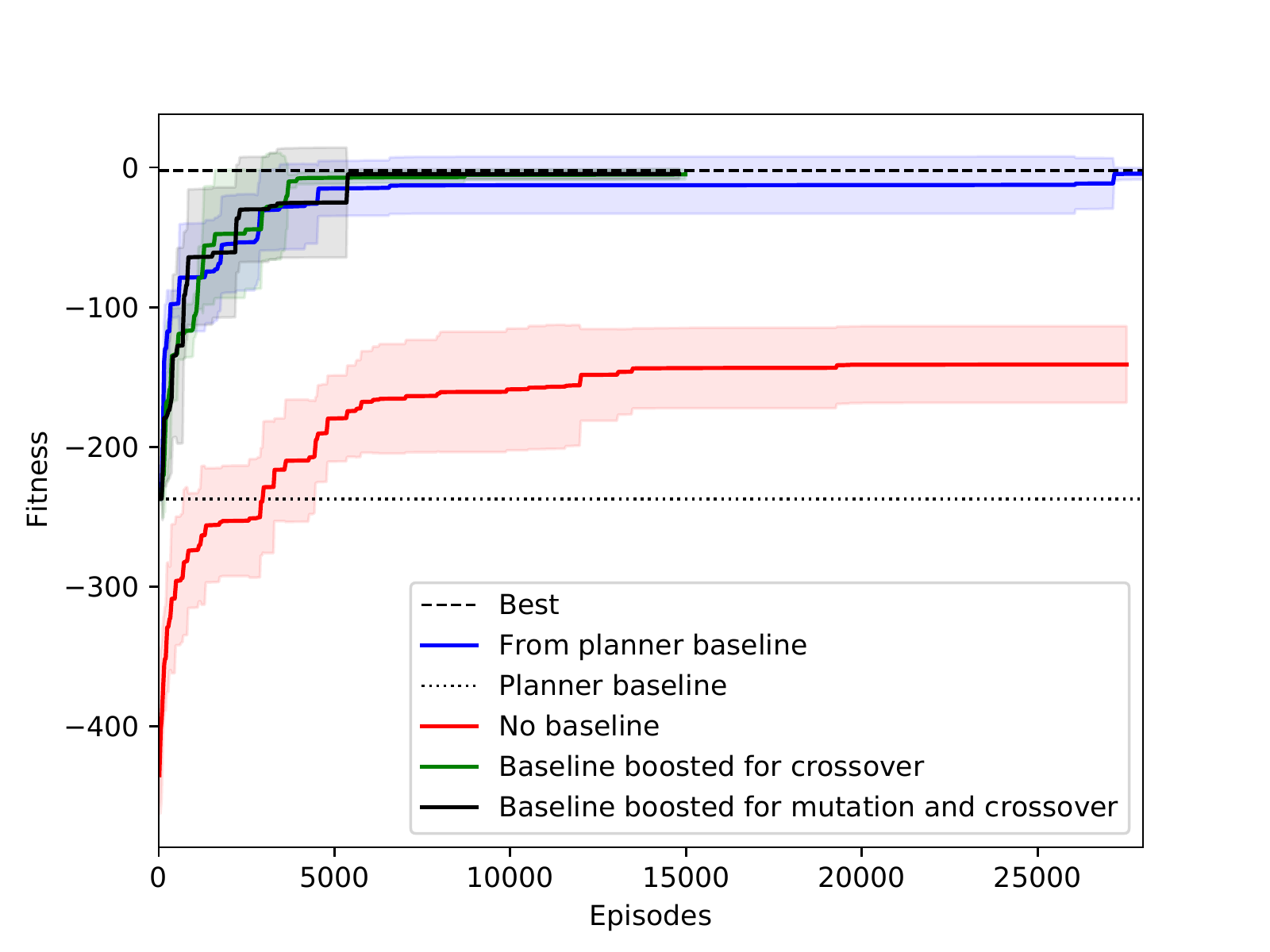}}
\caption{Learning curves for task 2.}
\label{fig:cro_learning_curve}
\end{figure}

Figure \ref{fig:cro_learning_curve} shows the learning curves for task 2 with the mean and standard deviation of 5 runs for each variant. It is clear that using the baseline not only allows the GP to start from a higher level, it also learns faster from that point as it can reuse subtrees from the planner baseline.
\par
\begin{figure}[htbp]
\centerline{\includegraphics[width=0.5\textwidth]{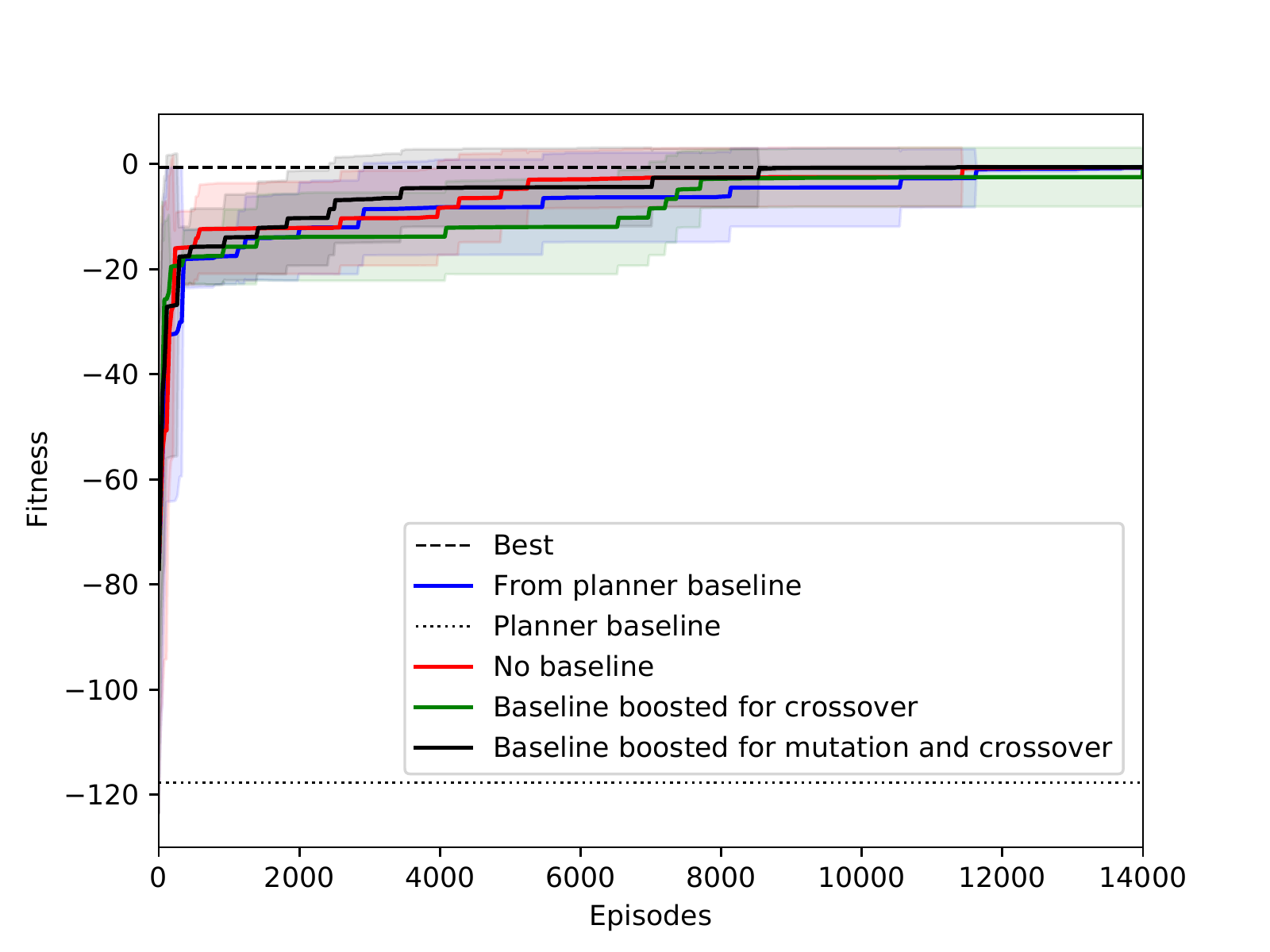}}
\caption{Learning curves for task 3.}
\label{fig:balance_learning_curve}
\end{figure}
\begin{figure}[htbp]
\centering
\centerline{\includegraphics[width=0.5\textwidth]{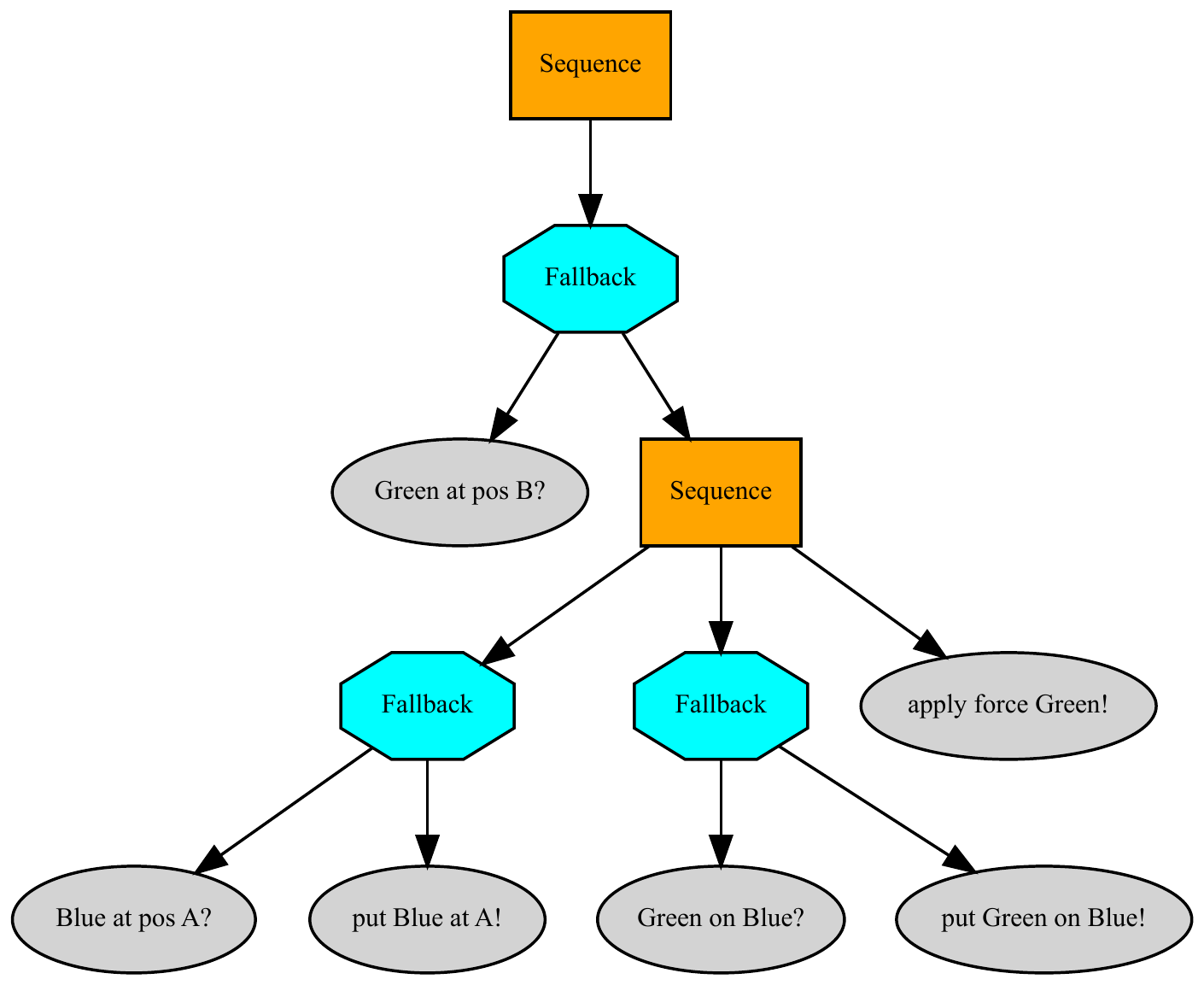}}
\caption{Planned BT for task 3.}
\label{fig:balance_bt_planned}
\end{figure}
\begin{figure}[htbp]
\centering
\centerline{\includegraphics[width=0.5\textwidth]{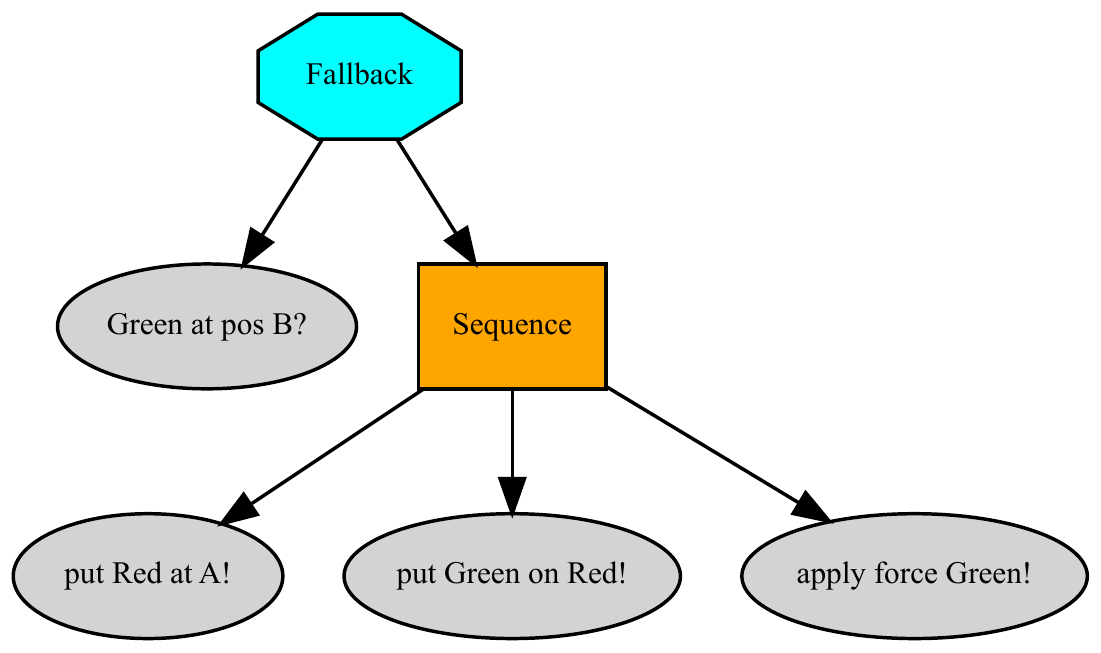}}
\caption{Converged BT for task 3.}
\label{fig:balance_bt_solved}
\end{figure}

Figure \ref{fig:balance_learning_curve} shows the learning curves for task 3 with the mean and standard deviation of 10 runs for each variant. In this experiment, all runs find a local minimum around \mbox{-19} very early  which is not too surprising as it consists of moving the green brick to below its target position and can be solved with only one behavior. Getting out of the local minimum requires a whole subtree that puts the blue brick as base and green brick on top, which takes a lot more time since there are no intermediate solutions that outperform the local minimum. For this task, there is no significant difference in the performance between the methods. This is not surprising when looking at the planned BT (Figure \ref{fig:balance_bt_planned}) vs a converged BT (Figure \ref{fig:balance_bt_solved}). Only a few single behaviors of the planned BT appear in the converged solution.


\par
\begin{figure}[htbp]
\centerline{\includegraphics[width=0.5\textwidth]{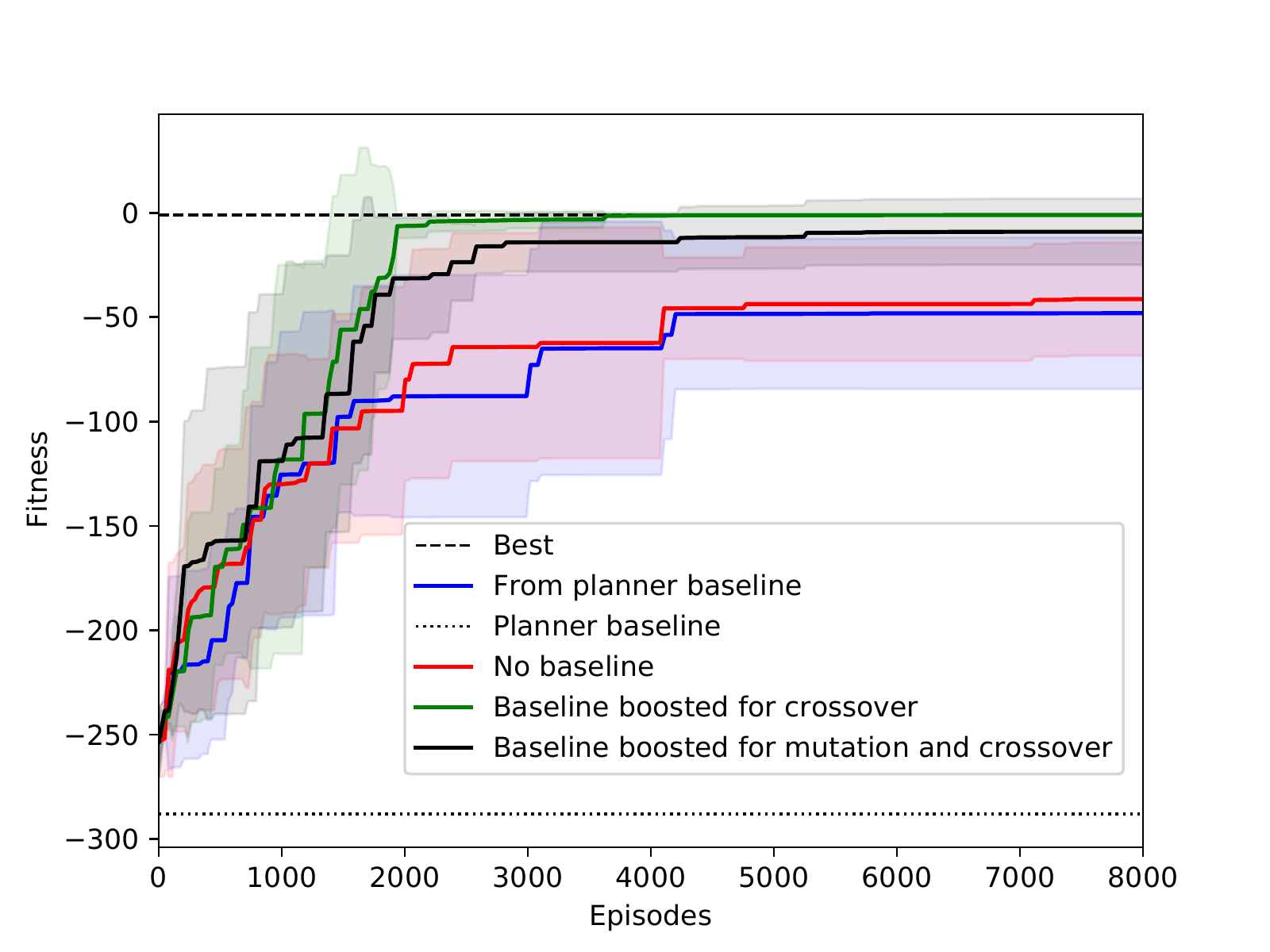}}
\caption{Learning curves for task 4.}
\label{fig:blocking_learning_curve}
\end{figure}
\begin{figure}[htbp]
\centerline{\includegraphics[width=0.5\textwidth]{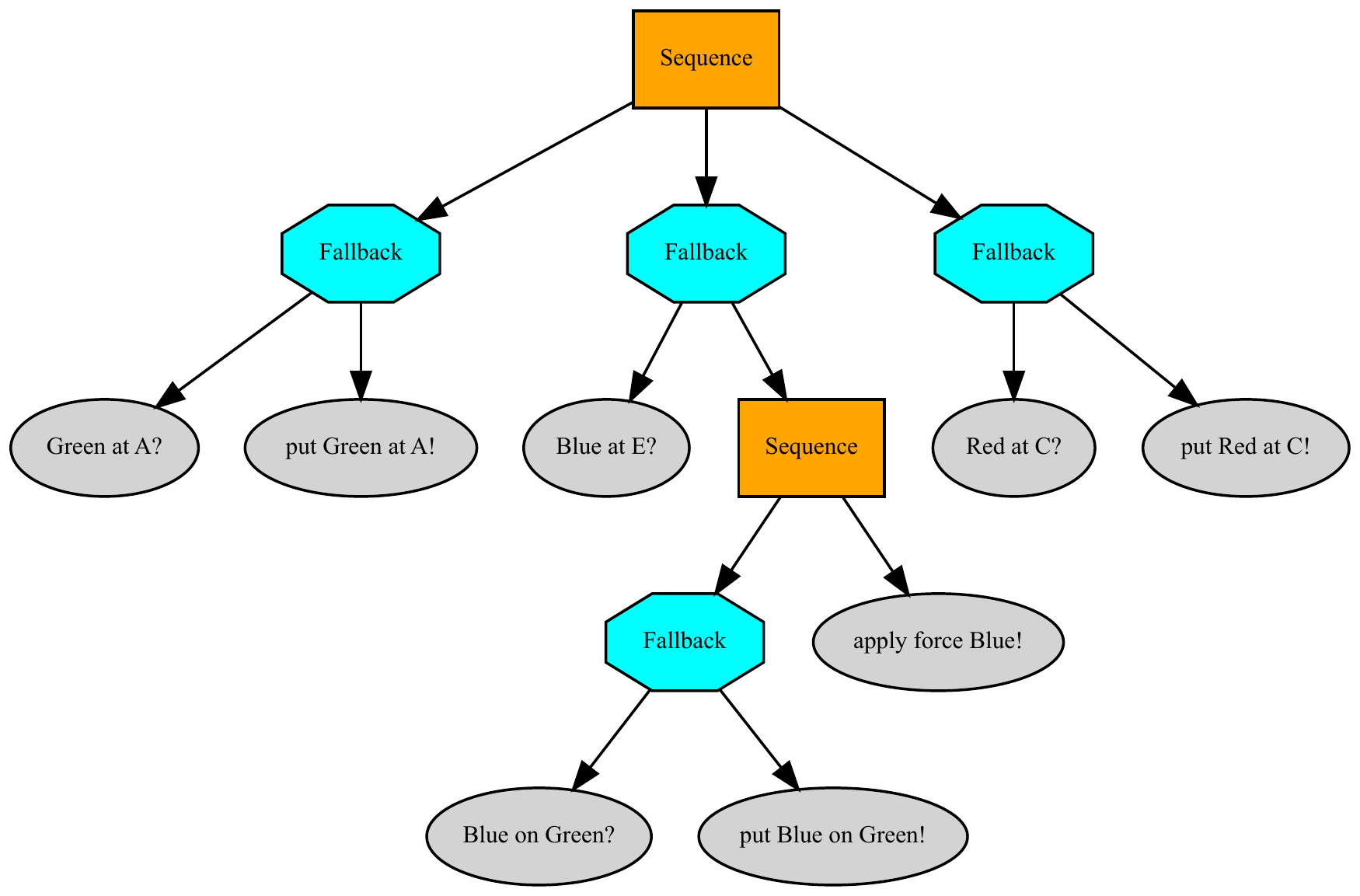}}
\caption{Planned BT for task 4.}
\label{fig:blocking_bt_planned}
\end{figure}
\begin{figure}[htbp]
\centerline{\includegraphics[width=0.5\textwidth]{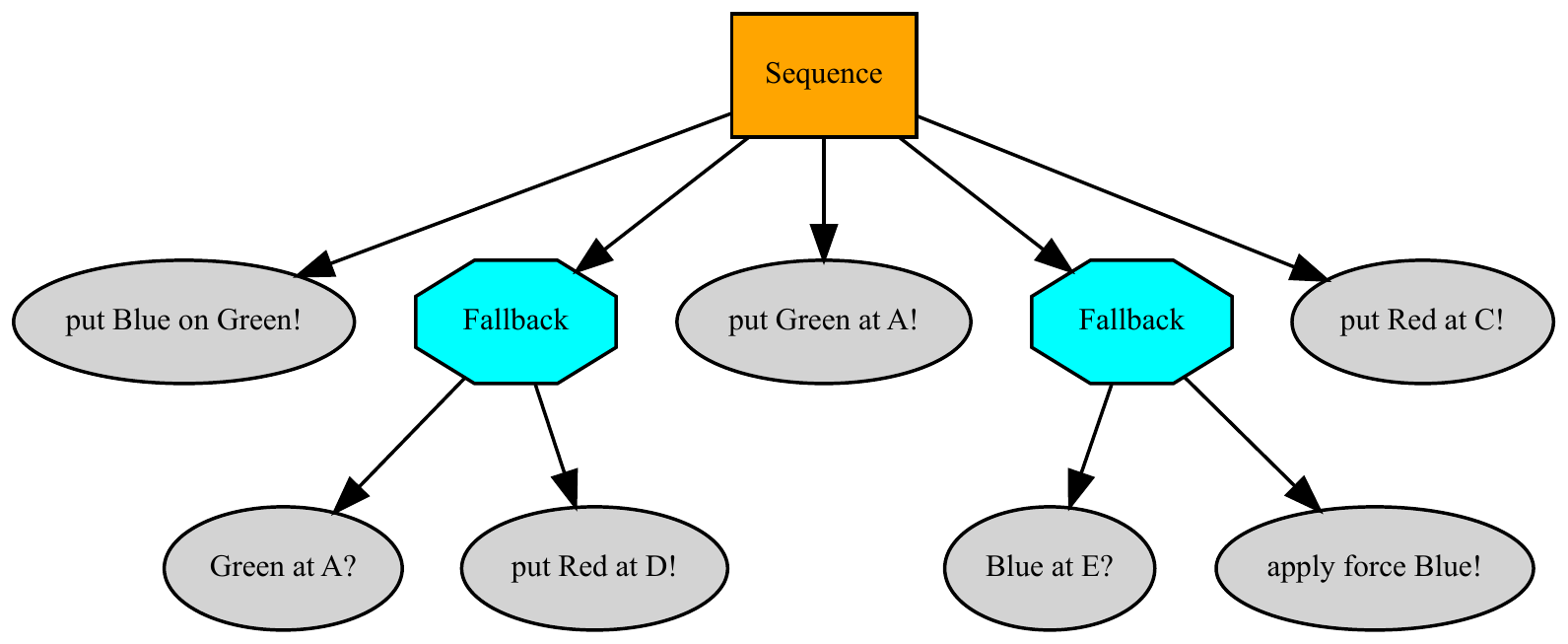}}
\caption{Converged BT for task 4.}
\label{fig:blocking_bt_solved}
\end{figure}

Figure \ref{fig:blocking_learning_curve} shows the learning curves for task 4 with the mean and standard deviation of 5 runs for each variant. Even without baseline the GP algorithm quickly finds a solution that is at least as good as the planner baseline. However, when boosted the planned BT helps speed up the convergence rate.
Figure \ref{fig:blocking_bt_planned} shows the behavior tree from the planner for task 4 and Figure \ref{fig:blocking_bt_solved} shows an example converged solution from the GP. Notably, besides changing the order of the behaviors and adding an intermediate position, the GP has also removed a number of redundant conditions to make the tree smaller. The positions in the behaviors correspond to the positions marked in Figure \ref{fig:blocking} and E is the position elevated one brick height above A.

\par
Interesting to note is that given the parameters of Table \ref{GP pars}, a maximum of 32 new BTs can be discovered each new generation. Given how many episodes were in the end run for each experiment, the number of episodes by unique BTs were generally at over 90\% of the maximum possible for the number of generations, meaning that very few trees were encountered more than once. Typically the final trees had around 10 nodes. With approximately 30 different nodes to search from, this means that the number of possible trees of just that size are in the order of $30^{10}$. Without doing any exact calculations, given that the total search space also consists of trees of other sizes it seems that even this very simple algorithm is in fact quite efficient given that it converges reliably in around 10 000 episodes, depending on the task.

\subsection{Experiments on real robot}
\begin{figure}[htbp]
\centerline{\includegraphics[width=0.5\textwidth]{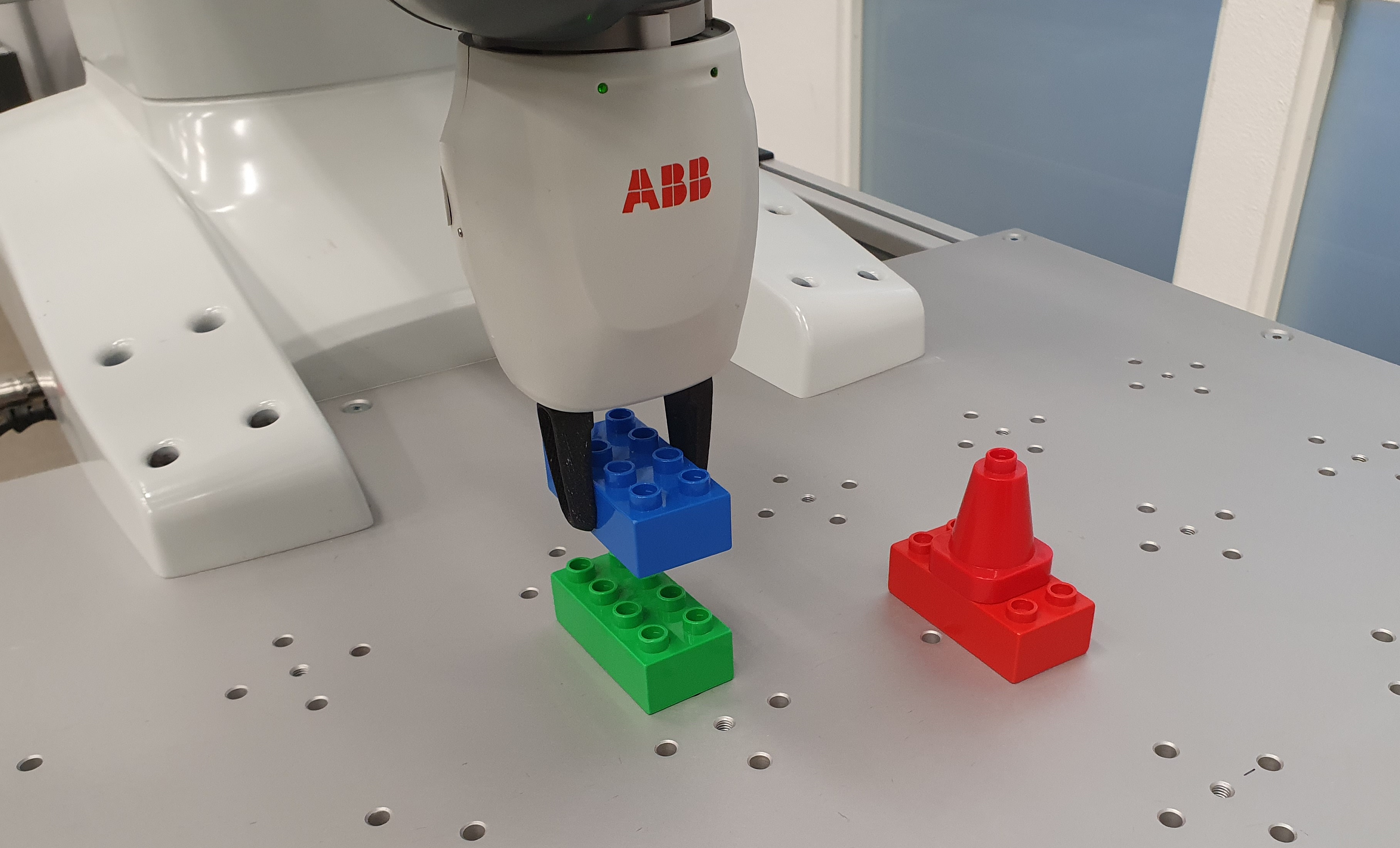}}
\caption{ABB YuMi robot performing task 4.}
\label{fig:real}
\end{figure}

We ran the solutions with the high level behaviors on a real ABB YuMi\textsuperscript{\textcopyright} robot to show that sim-to-real transfer is indeed possible; see Figure \ref{fig:real} and the video attached to the paper. For the purposes of this paper, we assumed in the simulations that the locations of the bricks were known. In reality this would need to be handled with a sensor system but as this was not within the scope of this paper we did not incorporate object detection in the real experiments. Instead we used hard coded assumptions of brick positions and the behaviors were translated to a language of the robots software. Therefore it was not exact copies of the behavior trees, but the movements of the robot gripper use the same target references as generated by the behavior trees in the simulations. One solution occasionally found in simulations was to drop the bricks onto the table repeatedly instead of pressing them together. While it turns out that this works also in reality, it is not very reliable as the bricks will sometimes bounce away or break apart. The simulations will also give very different results for slightly different initial states so this could be avoided with traditional sim-to-real methods such as domain randomization but that is left to future work. 

\section{Conclusions}
In conclusion, the results confirm that the proposed method of combining planning and learning perform well on different types of problems. Planners perform well when they have access to the necessary knowledge but fail when that is missing. Learning algorithms are slower but can learn the properties of the environment in which the agent acts. Combining planning and learning can yield an algorithm that at times outperforms both of them used separately and never performs worse for any of the evaluation tasks.
Our contributions are a proposed method to combine planning and learning and showing that this enables learning of behavior trees to be done more efficiently for a variety of robotic assembly problems. We showed that this type of high level learning of composites can be easily transferred to a real system without extensive further training.

\section{Future work}
While the tasks shown in our experiments were solved by the proposed algorithm with a relatively small computational effort, they were highly simplified with a very limited discrete parametrization of behaviors. Any complete solution would need to handle a much larger search space of possible behaviors as well as a much larger final program. The search space of possible BTs grows much faster than linearly with an increased number of nodes and behaviors which needs to be mitigated. One direction could be to try to automatically divide the full tasks into sub-tasks and solve them one by one. Building on the work in this paper, it could easily be improved with faster and more capable planners and more efficient learning algorithms. Further work must also be done to determine where and how exactly to best place the interface between planning and learning. Finally, we believe it should also be possible to draw inspiration from the recent results in reinforcement learning for neural networks and the methods used there, for example function approximation, and find ways to use similar methods when learning behavior trees.


\bibliographystyle{IEEEtran}
\bibliography{IEEEabrv,biblio}

\end{document}